\documentclass[runningheads]{llncs}
\pdfoutput=1 % arXiv: force pdfLaTeX (PDF figures)

\usepackage[year=2026]{eccv}
\usepackage{eccvabbrv}
\usepackage{graphicx}
\usepackage{booktabs}
\usepackage{amsmath,amssymb}
\usepackage{multirow}
\usepackage[accsupp]{axessibility}
\usepackage[a4paper,margin=30mm]{geometry}  % arXiv: widen LNCS text block (122mm -> 150mm)
\usepackage[breaklinks,colorlinks,citecolor=eccvblue]{hyperref}
\usepackage{orcidlink}
\usepackage{subcaption}
\usepackage{placeins}
\begin{document}

\title{Multi-Modal Conditioned High-Resolution Transformer for Urban Electromagnetic Field Map Prediction}
\titlerunning{Multi-Conditioned HRFormer for EMF Prediction}

\author{Do-Eon Kim\inst{1} \and
Dongryul Park\inst{2} \and
Seungyoung Ahn\inst{2} \and
Namwoo Kang\inst{2} \and
Seong-heum Kim\inst{3} \and
Seongsin Kim\inst{3}}
\authorrunning{D.-E. Kim et al.}
\institute{Soongsil University, Seoul 06978, Republic of Korea \and
Cho Chun Shik Graduate School of Mobility, Korea Advanced Institute of Science and Technology, Daejeon 34051, Republic of Korea \and
Department of Intelligent Semiconductors, Soongsil University, Seoul 06978, Republic of Korea}

\maketitle

% ==============================================================================
% ABSTRACT
% ==============================================================================
\begin{abstract}
Predicting electromagnetic field (EMF) strength in urban environments is essential for cellular network planning but computationally expensive with physics-based simulators.
We propose a multi-conditioned dense prediction framework that generates 500$\times$500 EMF maps from building layout images and antenna configurations.
Our architecture uses a High-Resolution Transformer (HRFormer) backbone with two complementary conditioning mechanisms: Feature-wise Linear Modulation (FiLM) injects scalar antenna parameters into all backbone stages, while cross-attention fuses 1-D radiation pattern tokens with spatial features at the deepest stage.
We further introduce transmitter-relative spatial channels encoding distance, proximity, and bearing from the antenna, enabling coordinate-consistent test-time augmentation (TTA) that reduces test MAE by 6.3\%.
To address the prediction difficulty imbalance across EMF maps, we design a composite loss combining masked L1, multi-scale structural similarity (MS-SSIM), and a focal L1 term that upweights high-signal pixels, outperforming individual loss components in all metrics.
Our best model achieves a test MAE of 0.0461, a 25.2\% improvement over a plain UNet baseline and 31.8\% over an HRFormer-only baseline.
\end{abstract}

% ==============================================================================
% 1. INTRODUCTION
% ==============================================================================
\section{Introduction}
\label{sec:intro}

The deployment and optimization of cellular networks requires accurate estimation of electromagnetic field (EMF) strength across urban areas.
Traditional approaches rely on physics-based ray-tracing simulators~\cite{3gpp38901} that, while accurate, are computationally prohibitive for large-scale network planning involving thousands of candidate antenna sites.
This motivates the development of fast, learned surrogates for EMF map prediction.

Recent deep learning approaches have shown promise for radio map prediction.
CNN-based methods~\cite{levie2021radiounet,ratnam2020fadenet} treat the task as image-to-image translation from building layouts to signal strength maps.
However, these methods typically concatenate all inputs as image channels, limiting their ability to exploit the distinct nature of different input modalities: the spatial building layout, scalar antenna parameters (position, height, power, tilt, azimuth), and angular radiation patterns.

We propose a multi-conditioned dense prediction framework that processes each modality through a dedicated pathway.
Our key insight is that antenna parameters and radiation patterns should \emph{condition} the visual feature extraction rather than merely augmenting the input.
Specifically, we employ two conditioning mechanisms atop a high-resolution transformer backbone:

\begin{enumerate}
    \item \textbf{FiLM conditioning}~\cite{perez2018film} applies learned affine transforms derived from scalar antenna parameters at every backbone stage, modulating feature responses based on transmission power, antenna height, and beam direction.
    \item \textbf{Cross-attention}~\cite{vaswani2017attention} enables the deepest backbone features to attend to tokenized radiation patterns, allowing the model to learn antenna-specific propagation characteristics.
\end{enumerate}

Additionally, we introduce \emph{transmitter spatial channels} that encode the geometric relationship between each pixel and the antenna position, providing explicit distance and bearing information.
These channels enable a coordinate-consistent test-time augmentation (TTA) strategy where spatial channels are recomputed for each augmentation transform rather than naively flipped, yielding a 6.3\% MAE reduction.

We validate our approach through controlled ablation studies on a dataset of 768 simulated urban EMF maps.
Our best model achieves a test MAE of 0.0461, a 25.2\% improvement over a plain UNet baseline and 31.8\% over an HRFormer-only baseline.

Our contributions are as follows:
\begin{itemize}
    \item A multi-modal conditioning architecture integrating FiLM, cross-attention, and transmitter spatial channels for EMF map prediction.
    \item A composite loss combining masked L1, MS-SSIM~\cite{wang2003msssim}, and a focal L1 term~\cite{lin2017focal} that upweights high-signal regions, addressing prediction difficulty imbalance across EMF maps.
    \item Coordinate-consistent TTA that properly recomputes spatial channels under geometric transforms, achieving 6.3\% MAE reduction.
    \item Controlled experiments that quantify the contribution of each component and reveal that conditioning mechanism design matters more than backbone scale alone.
\end{itemize}

% ==============================================================================
% 2. RELATED WORK
% ==============================================================================
\section{Related Work}
\label{sec:related}

\noindent\textbf{Conventional EMF evaluation.}
International standards for base station EMF compliance rely on calculation-based methods using free-space path loss (FSPL) models~\cite{ituk70,ituk100}.
While simple, these methods are limited to line-of-sight (LoS) conditions and worst-case scenarios, making them inapplicable in complex urban environments.
Simulation-based alternatives solve Maxwell's equations~\cite{lodato2024raytracing,lala2018fdtd} for higher accuracy, but require detailed 3-D models and prohibitive computation time.
Measurement-based approaches~\cite{kapetanakis2022assessment} provide reliable evaluations in actual environments; however, 5G densification and MIMO beam configurations drastically increase the required number of measurement points, making this approach increasingly impractical.

\noindent\textbf{AI-based radio map prediction.}
Early AI approaches predict path loss from vehicle-based measurements and satellite images~\cite{thrane2020pathloss,cheng2020cnn_mmwave}, but limited spatial coverage restricts generalization.
To overcome this, CNN-based methods~\cite{levie2021radiounet,zhang2020cellular,krijestorac2023agile,lee2024scalable} take building layouts and antenna locations as spatial inputs and predict path loss maps over entire regions.
Recent work extends this to transformer~\cite{liu2025deformable} and Mamba architectures~\cite{jia2025radiomamba}.
However, most approaches assume isotropic transmitters~\cite{yapar2022radiomap_dataset} and concatenate all inputs as image channels, failing to account for directional antenna characteristics.

\noindent\textbf{EMF prediction with directional antennas.}
Park~\etal~\cite{park2024emf_hotspot} propose a physics-informed input encoding for directional base station EMF prediction using a UNet.
Kim~\etal~\cite{kim2025emf_vit} evaluate various ViT architectures on the same dataset.
However, both rely on standard architectures without task-specific conditioning: antenna parameters are concatenated as input channels rather than used to modulate intermediate features.
Our work introduces structured multi-modal conditioning---FiLM for scalar antenna parameters and cross-attention for radiation patterns---that explicitly encodes antenna characteristics into the feature extraction process, yielding substantial accuracy gains over the concatenation-based approach.

\noindent\textbf{Multi-modal conditioning in vision.}
FiLM~\cite{perez2018film} conditions visual features via learned affine transforms.
Cross-attention~\cite{vaswani2017attention} provides a richer pathway by allowing spatial features to attend to external tokens.
We combine both to condition on structurally different modalities: scalar parameters via FiLM and angular radiation patterns via cross-attention, atop an HRFormer~\cite{yuan2021hrformer} backbone that maintains multi-resolution features throughout~\cite{sun2019hrnet}.

% ==============================================================================
% 3. METHOD
% ==============================================================================
\section{Method}
\label{sec:method}

\begin{figure}[t]
    \centering
    \includegraphics[width=\linewidth]{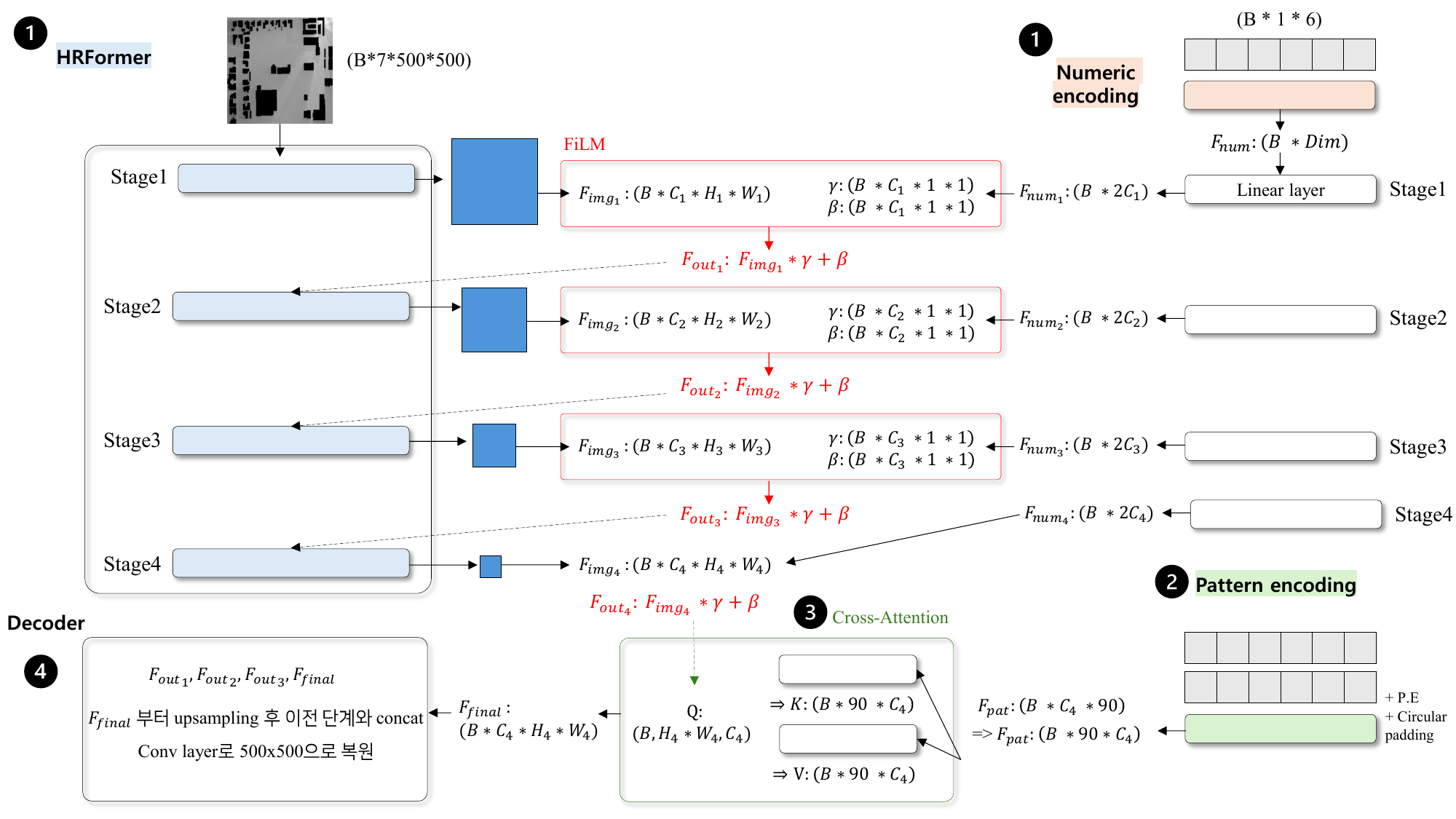}
    \caption{Overview of the proposed framework. (a)~A 7-channel input (building layout + Tx spatial channels) is processed by the HRFormer backbone. (b)~FiLM injects scalar antenna parameters ($\boldsymbol{\gamma}, \boldsymbol{\beta}$) at all four stages. (c)~Cross-attention at Stage~4 fuses spatial features with 90 pattern tokens from 360$^\circ$ radiation patterns. (d)~UNet decoder with skip connections produces the $500{\times}500$ EMF map.}
    \label{fig:architecture}
\end{figure}

\cref{fig:architecture} overviews our framework.
A 7-channel input feeds the HRFormer backbone, producing multi-scale features at four resolutions.
FiLM modulates all stages using scalar antenna parameters; cross-attention at the deepest stage fuses spatial features with radiation pattern tokens.
A UNet decoder upsamples conditioned features to produce the EMF map.

\subsection{Problem Formulation}
\label{sec:problem}

Given a 2-D building layout $B \in \mathbb{R}^{H \times W}$, antenna parameters $\mathbf{a} {=} [x, y, z, \delta, p, \phi] \in \mathbb{R}^{6}$ (position, height, downtilt, power, azimuth), and radiation patterns $\mathbf{r}_h, \mathbf{r}_v \in \mathbb{R}^{360}$, we predict a dense EMF map $\hat{\mathbf{y}} \in \mathbb{R}^{H \times W}$ by learning $f_\theta\!: (B, \mathbf{a}, \mathbf{r}_h, \mathbf{r}_v) \mapsto \hat{\mathbf{y}}$.

\subsection{Input Representation}
\label{sec:input}

\begin{figure}[t]
    \centering
    \includegraphics[width=\linewidth]{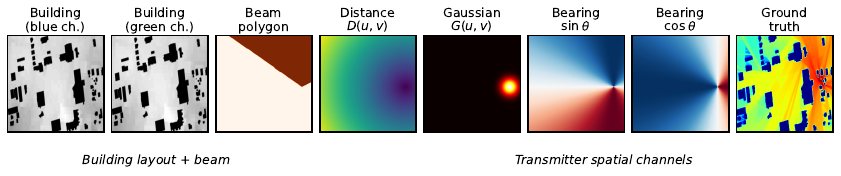}
    \caption{Input channel visualization. The first three channels encode building geometry and beam coverage; the remaining four encode transmitter-relative spatial priors (distance, proximity, and directional bearing).}
    \label{fig:input_channels}
\end{figure}

The network input $X_0 \in \mathbb{R}^{H \times W \times 7}$ concatenates the building layout with transmitter-relative spatial channels.
The first two channels encode building geometry (exterior walls and interior regions), the third encodes a beam coverage polygon constructed from the antenna position and azimuth angle, indicating the approximate main-lobe coverage sector.
The remaining four encode the relationship between pixel $(u,v)$ and transmitter $(u_t, v_t)$.
Let $\Delta u = u{-}u_t$, $\Delta v = v{-}v_t$:
\begin{align}
    D &= \tfrac{\sqrt{\Delta u^2 + \Delta v^2}}{d_{\max}}, &
    G &= \exp\!\bigl({-}\tfrac{\Delta u^2 + \Delta v^2}{2\sigma^2}\bigr), \label{eq:dist} \\
    S &= \sin\!\bigl(\mathrm{atan2}(\Delta v, \Delta u)\bigr), &
    C &= \cos\!\bigl(\mathrm{atan2}(\Delta v, \Delta u)\bigr), \label{eq:sin}
\end{align}
where $d_{\max} = H\sqrt{2}$ normalizes distance to $[0,1]$ and $\sigma{=}30$ controls the Gaussian proximity width.
These channels provide explicit spatial priors about signal attenuation, antenna proximity, and propagation direction.
\cref{fig:input_channels} visualizes all seven channels for a representative sample.

\subsection{Backbone}
\label{sec:backbone}

We use HRFormer~\cite{yuan2021hrformer}, which maintains four parallel branches at progressively lower resolutions.
For $500{\times}500$ input, the branches produce features at resolutions $H/4$ through $H/32$ with dimensions $(C_1,\ldots,C_4)$.
We evaluate Small (32--256) and Base (78--624) channel variants.

\subsection{FiLM Conditioning}
\label{sec:film}

To condition feature extraction on antenna parameters, we apply Feature-wise Linear Modulation (FiLM)~\cite{perez2018film} at each backbone stage.
A two-layer MLP $g_s$ maps the antenna parameter vector $\mathbf{a} \in \mathbb{R}^6$ to per-channel scale and shift parameters:
\begin{equation}
    (\boldsymbol{\gamma}_s, \boldsymbol{\beta}_s) = g_s(\mathbf{a}), \quad
    \tilde{F}_s = \boldsymbol{\gamma}_s \odot F_s + \boldsymbol{\beta}_s,
    \label{eq:film}
\end{equation}
where $F_s \in \mathbb{R}^{N_s \times C_s}$ is the feature map at stage $s$ and $\odot$ denotes channel-wise multiplication.
FiLM is applied at all four stages ($s \in \{1,2,3,4\}$), allowing the network to adapt feature responses from early texture-level to late semantic-level representations based on transmission parameters.

\subsection{Radiation Pattern Cross-Attention}
\label{sec:crossattn}

Radiation patterns encode antenna-specific directional gain characteristics that cannot be adequately captured by scalar conditioning alone.
We first tokenize the 360$^\circ$ pattern into $M=90$ tokens using a 1-D CNN encoder $E_p$ that groups every 4 degrees:
\begin{equation}
    T = E_p(\mathbf{r}) \in \mathbb{R}^{M \times d_p},
\end{equation}
where $\mathbf{r} = [\mathbf{r}_h; \mathbf{r}_v] \in \mathbb{R}^{720}$ concatenates both patterns and $d_p = C_4$ matches the deepest stage channel dimension.

At stage 4, spatial features serve as queries and pattern tokens as keys/values in a standard transformer decoder:
\begin{equation}
    Q = XW_Q, \; K = TW_K, \; V = TW_V, \;
    X' = X + \mathrm{softmax}\!\left(\frac{QK^\top}{\sqrt{d_k}}\right)V,
    \label{eq:crossattn}
\end{equation}
where $X \in \mathbb{R}^{N_4 \times C_4}$ with $N_4 = 256$ spatial tokens at $16 \times 16$ resolution.
We use 8 attention heads and 2 transformer decoder layers.

\subsection{Decoder}
\label{sec:decoder}

A UNet-style decoder~\cite{ronneberger2015unet} progressively upsamples the conditioned features using skip connections from the backbone.
Five decoder stages progressively recover the full spatial resolution: three stages fuse upsampled features with backbone skip connections from stages 3, 2, and 1, while two additional stages upsample without skip connections to reach the target resolution.
The final $1{\times}1$ convolution with ReLU activation produces the predicted EMF map $\hat{\mathbf{y}} \in \mathbb{R}^{500 \times 500}$ (center-cropped from the padded backbone output).

\subsection{Loss Function}
\label{sec:loss}

We use a composite loss combining three complementary objectives:
\begin{equation}
    \mathcal{L} = \mathcal{L}_{\ell_1} + \lambda_{\text{ms}} \mathcal{L}_{\text{MS-SSIM}} + \lambda_f \mathcal{L}_{\text{Focal-}\ell_1},
    \label{eq:loss}
\end{equation}
with $\lambda_{\text{ms}} = 2.0$ and $\lambda_f = 0.25$.

The masked L1 loss operates only on outdoor pixels $\Omega$ (excluding building interiors):
\begin{equation}
    \mathcal{L}_{\ell_1} = \frac{1}{|\Omega|} \sum_{u \in \Omega} |\hat{y}_u - y_u|.
\end{equation}

The MS-SSIM loss~\cite{wang2003msssim} captures perceptual structural similarity at multiple scales.
Inspired by the focal weighting idea of~\cite{lin2017focal}, our Focal L1 loss reweights pixel errors to emphasize high-signal regions where prediction accuracy is most critical:
\begin{equation}
    \mathcal{L}_{\text{Focal-}\ell_1} = \frac{1}{|\Omega|} \sum_{u \in \Omega} w_u \cdot |\hat{y}_u - y_u|, \quad
    w_u = \gamma \!\cdot\! \bigl(y_u / y_{\max}\bigr)^q + 1,
\end{equation}
where $q{=}0.5$ controls the focusing strength, $\gamma{=}2.0$ scales the dynamic range of the weights, and $y_{\max}$ normalizes targets to $[0,1]$.
The additive constant ensures a minimum weight of 1 for all pixels.

\subsection{Coordinate-Consistent Test-Time Augmentation}
\label{sec:tta}

Standard flip-based TTA naively transforms all input channels identically.
However, our transmitter spatial channels (\cref{sec:input}) encode absolute geometric relationships that must be recomputed---not simply flipped---under spatial transforms.

For each augmentation $t \in \mathcal{T} = \{\mathrm{Id}, \mathrm{H\text{-}flip}, \mathrm{V\text{-}flip}, \mathrm{HV\text{-}flip}\}$, we:
(1)~transform the building layout $B^{(t)} = t(B)$,
(2)~transform the antenna coordinates $\mathbf{a}^{(t)}$ (flipping $x \to W{-}x$ or $y \to H{-}y$ and adjusting azimuth),
(3)~\emph{recompute} spatial channels $D^{(t)}, G^{(t)}, S^{(t)}, C^{(t)}$ from the transformed coordinates.
The final prediction averages over inverse-transformed outputs:
\begin{equation}
    \hat{\mathbf{y}}_{\text{TTA}} = \frac{1}{|\mathcal{T}|} \sum_{t \in \mathcal{T}} t^{-1}\!\left(\hat{\mathbf{y}}^{(t)}\right).
    \label{eq:tta}
\end{equation}

% ==============================================================================
% 4. EXPERIMENTS
% ==============================================================================
\section{Experiments}
\label{sec:exp}

\subsection{Dataset and Setup}
\label{sec:dataset}

We use a dataset of 768 simulated EMF maps at 3.5\,GHz (5G NR n78 band) generated by a commercial radio propagation simulator.
Each sample consists of a $500 \times 500$ pixel urban area ($\sim$1\,m/pixel) containing a building layout image, antenna configuration parameters, radiation pattern files, and the corresponding ground-truth EMF strength map.
The dataset covers 256 distinct building layouts with 3 antenna configurations each (varying azimuth, downtilt, and power).

We split the 768 samples randomly (seed=42) into 614 training, 76 validation, and 78 test samples.
Note that the same building layout may appear in multiple splits with different antenna configurations; this reflects realistic deployment where the same area is served by different base stations.
Training data is augmented with horizontal, vertical, and combined flips (4$\times$), yielding 2{,}456 effective training samples.
Augmented samples have their antenna coordinates and azimuth angles transformed consistently.

All models are trained for 200 epochs with AdamW~\cite{loshchilov2019adamw} optimizer, linear warmup (5 epochs) followed by cosine annealing, batch size of 4 (or 8 where noted), and gradient clipping at 1.0.
Mixed precision training (FP16) is used where numerically stable.
EMF maps are normalized to $[0,1]$ (mapping simulator output to grayscale intensity).
The primary evaluation metric is Mean Absolute Error (MAE) computed only on outdoor pixels (excluding building interiors).
We additionally report RMSE, SSIM, and Hotspot IoU (intersection-over-union of the top-5\% signal strength regions).

\subsection{Baseline Models}
\label{sec:baselines}

We compare against two baselines representing standard approaches:

\noindent\textbf{Plain UNet.}
A standard 4-level UNet encoder-decoder~\cite{ronneberger2015unet} with no conditioning mechanisms.
Antenna parameters and radiation patterns are converted to spatial maps via MLPs and concatenated with the building layout (6 channels total, 35.0M parameters).
This follows the architecture of prior CNN-based EMF prediction work~\cite{park2024emf_hotspot}.

\noindent\textbf{HRFormer + Regressor.}
An HRFormer-Small backbone with a simple progressive-upsampling regressor~\cite{kim2025emf_vit}.
The same 6-channel input concatenation strategy is used.
Only the finest-resolution branch output is passed to the regressor (8.4M parameters).
Neither FiLM conditioning nor cross-attention is applied.

Both baselines use L1 loss and identical data splits for fair comparison.

\subsection{Main Results}
\label{sec:results}

\cref{tab:main} presents the test set evaluation across all key model variants.
Our best single-model result achieves a test MAE of 0.0461, representing a 25.2\% reduction from the UNet baseline and 31.8\% from the HRFormer baseline.
SSIM improves from 0.918/0.891 to 0.949, indicating better structural fidelity in the predicted EMF maps.

\begin{table}[t]
    \centering
    \caption{Test set results on 78 samples. All metrics computed on outdoor pixels only. Backbone: S = HRFormer-Small, B = HRFormer-Base. Best in each column is \textbf{bold}.}
    \label{tab:main}
    \begin{tabular}{llcccccc}
        \toprule
        Method & Conditioning & Backbone & MAE$\downarrow$ & RMSE$\downarrow$ & SSIM$\uparrow$ & IoU$\uparrow$ & ms \\
        \midrule
        Plain UNet & None & UNet & 0.0616 & 0.0877 & 0.9175 & 0.326 & 7.6 \\
        HRFormer+Reg & None & S & 0.0676 & 0.1038 & 0.8907 & 0.300 & 15.4 \\
        \midrule
        Ours (L1 only) & FiLM+CA & S & 0.0606 & 0.0920 & 0.9069 & 0.412 & --- \\
        Ours & FiLM(1-3)+CA(4) & S & 0.0546 & 0.0748 & 0.9388 & \textbf{0.423} & --- \\
        Ours & FiLM(all)+CA(4) & S & 0.0533 & 0.0722 & 0.9425 & 0.406 & --- \\
        Ours & FiLM(all)+CA(4) & B & 0.0492 & \textbf{0.0665} & \textbf{0.9490} & 0.413 & 19.8 \\
        Ours (best) & FiLM(all)+CA(4) & B & \textbf{0.0461} & 0.0670 & 0.9480 & 0.406 & 19.8 \\
        \midrule
        Ours + TTA & FiLM(all)+CA(4) & B & \textbf{0.0461} & --- & --- & --- & 79.2 \\
        \bottomrule
    \end{tabular}
\end{table}

\noindent\textbf{Hotspot localization.}
Beyond pixel-level accuracy, the Hotspot IoU metric reveals how well each model localizes high-signal regions---a critical requirement for network coverage planning.
Baselines without conditioning (UNet: 0.326, HRFormer+Reg: 0.300) lag substantially behind all conditioned variants (0.406--0.423), demonstrating that FiLM and cross-attention help the model capture antenna-specific propagation hotspots.

\noindent\textbf{Inference speed.}
On an RTX 5090 GPU, our full model runs at 19.8\,ms per sample---roughly 50 predictions per second.
Even with 4-fold TTA ($\sim$80\,ms), inference remains orders of magnitude faster than physics-based simulation.
The UNet baseline is faster (7.6\,ms) but at significantly lower accuracy.

\subsection{Qualitative Results}
\label{sec:qualitative}

\begin{figure}[t]
    \centering
    \includegraphics[width=0.95\linewidth]{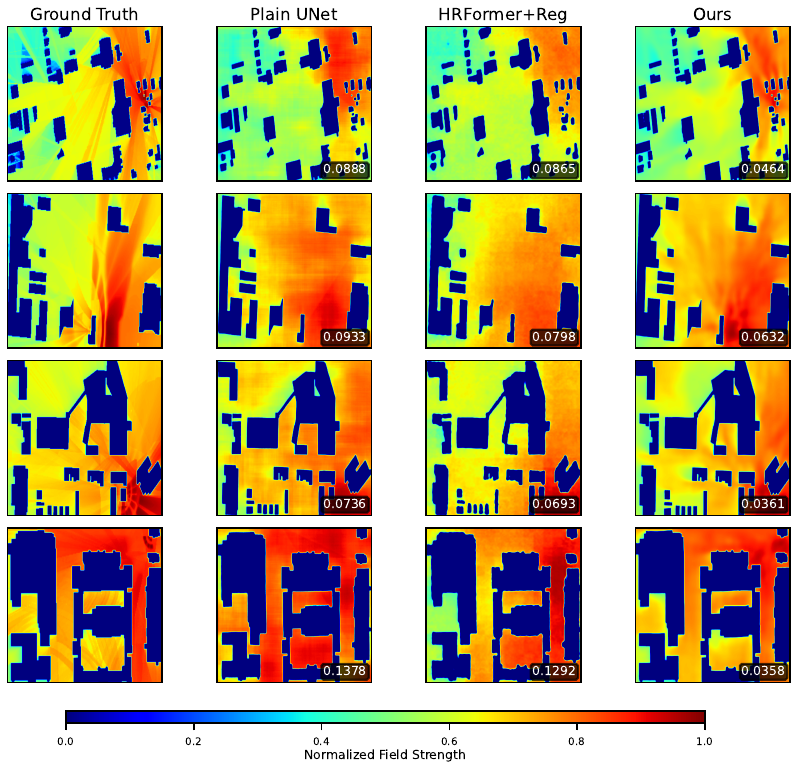}
    \caption{Prediction comparison across four test samples (rows) with different antenna azimuths. Columns: ground truth, plain UNet, HRFormer baseline (no conditioning), and our model. Per-sample MAE shown in corners.}
    \label{fig:qualitative}
\end{figure}

\cref{fig:qualitative} provides a side-by-side comparison of predictions from three models across four diverse test samples.
The plain UNet produces blurred predictions with substantial errors around building boundaries.
The HRFormer baseline (without conditioning) captures spatial structure but fails to adapt to antenna-specific propagation patterns.
Our conditioned model generates the sharpest predictions with errors concentrated only at building edges where diffraction effects are most complex.

\subsection{Ablation Studies}
\label{sec:ablation}

We conduct four rounds of ablation experiments, each building on the best configuration from the previous round.

\noindent\textbf{Round 1: Loss function} (\cref{tab:loss}).
Starting from the full conditioning architecture (FiLM + cross-attention, HRFormer-Small), we compare loss formulations.
The combined loss (L1 + MS-SSIM + Focal L1) consistently outperforms individual components, with constant weighting preferred over cosine scheduling.

\begin{table}[t]
    \centering
    \caption{Round 1 --- Loss ablation. All use HRFormer-Small with FiLM(1-3)+CA(4), batch size 4.}
    \label{tab:loss}
    \begin{tabular}{lc}
        \toprule
        Loss function & Val MAE$\downarrow$ \\
        \midrule
        L1 only & 0.062 \\
        L1 + Focal L1 (cosine schedule) & 0.060 \\
        L1 + MS-SSIM ($\lambda$=2.0, constant) & 0.059 \\
        \textbf{L1 + MS-SSIM + Focal (constant)} & \textbf{0.0575} \\
        L1 + MS-SSIM $\rightarrow$ Focal (cosine) & 0.060 \\
        \bottomrule
    \end{tabular}
\end{table}

\noindent\textbf{Round 2: Conditioning architecture} (\cref{tab:arch}).
Using the best loss, we vary FiLM and cross-attention stage assignments.
Extending FiLM to all four stages yields the best result, while adding cross-attention to stage 3 provides marginal improvement at the cost of numerical instability requiring FP32 training.

\begin{table}[t]
    \centering
    \caption{Round 2 --- Architecture ablation. Best loss from Round~1, batch size 8.}
    \label{tab:arch}
    \begin{tabular}{lc}
        \toprule
        Conditioning & Val MAE$\downarrow$ \\
        \midrule
        FiLM only (no CA, no spatial) & 0.065 \\
        FiLM(1,2,3) + CA(4) & 0.061 \\
        FiLM(1,2,3) + CA(3,4) & 0.060 \\
        \textbf{FiLM(1,2,3,4) + CA(4)} & \textbf{0.0588} \\
        FiLM(1,2,3,4) + CA(3,4) & 0.059 \\
        \bottomrule
    \end{tabular}
\end{table}

\noindent\textbf{Round 3: Backbone capacity} (\cref{tab:backbone}).
Scaling from HRFormer-Small to HRFormer-Base yields a substantial improvement (0.0588$\,{\to}\,$0.0559).
Applying masked loss causes training divergence due to numerical instability in the MS-SSIM computation on sparse masked regions, while reducing batch size from 8 to 4 slightly degrades performance.

\begin{table}[t]
    \centering
    \caption{Round 3 --- Backbone and training ablation. Based on best Round~2 configuration.}
    \label{tab:backbone}
    \begin{tabular}{lccc}
        \toprule
        Variant & Val MAE$\downarrow$ & Val RMSE & Val SSIM \\
        \midrule
        HRFormer-Small, bs=8 & 0.0588 & --- & --- \\
        \textbf{HRFormer-Base, bs=4} & \textbf{0.0559} & \textbf{0.0728} & \textbf{0.9409} \\
        + masked loss & 0.44 (failed) & --- & --- \\
        Small, bs=4 & 0.0568 & 0.0767 & 0.9330 \\
        \bottomrule
    \end{tabular}
\end{table}

\noindent\textbf{Round 4: Learning rate tuning} (\cref{tab:lr}).
Fine-tuning the learning rate and weight decay for the larger HRFormer-Base backbone yields the overall best configuration: $\text{lr}_\text{head}=3\!\times\!10^{-4}$, $\text{lr}_\text{backbone}=1\!\times\!10^{-5}$, weight decay 0.02.

\begin{table}[t]
    \centering
    \caption{Round 4 --- Learning rate tuning on HRFormer-Base.}
    \label{tab:lr}
    \begin{tabular}{cccc}
        \toprule
        $\text{lr}_\text{head}$ & $\text{lr}_\text{bb}$ & Weight decay & Val MAE$\downarrow$ \\
        \midrule
        $1\!\times\!10^{-4}$ & $1\!\times\!10^{-5}$ & 0.05 & 0.0559 \\
        $3\!\times\!10^{-4}$ & $1\!\times\!10^{-5}$ & 0.05 & 0.0572 \\
        $\boldsymbol{3\!\times\!10^{-4}}$ & $\boldsymbol{1\!\times\!10^{-5}}$ & \textbf{0.02} & \textbf{0.0548} \\
        $2\!\times\!10^{-4}$ & $5\!\times\!10^{-6}$ & 0.05 & 0.0550 \\
        \bottomrule
    \end{tabular}
\end{table}

\subsection{Test-Time Augmentation Analysis}
\label{sec:tta_results}

\cref{tab:tta} shows the effect of coordinate-consistent TTA.
Recomputing spatial channels from transformed antenna coordinates (rather than naively flipping them) is critical: our coordinate-consistent approach yields a 6.3\% MAE reduction for the HRFormer-Base model. Naive flipping, which applies geometric transforms to precomputed spatial channels without recomputing them from the transformed antenna position, produces inconsistent distance and bearing information that can degrade rather than improve predictions.

\begin{table}[t]
    \centering
    \caption{TTA results on the test set (78 samples). Coordinate-consistent TTA recomputes spatial channels for each augmentation.}
    \label{tab:tta}
    \begin{tabular}{lccc}
        \toprule
        Model & Baseline MAE & +TTA MAE & Improvement \\
        \midrule
        HRFormer-Base & 0.04919 & \textbf{0.04609} & $-$6.3\% \\
        Best config & 0.04996 & 0.04698 & $-$6.0\% \\
        HRFormer-Small & 0.05063 & 0.04881 & $-$3.6\% \\
        \bottomrule
    \end{tabular}
\end{table}

\subsection{Where Does Conditioning Help?}
\label{sec:analysis}

To understand whether our conditioning modules capture physically meaningful patterns, we visualize their internal representations and analyze where prediction errors are reduced.

\begin{figure}[t]
    \centering
    \includegraphics[width=0.95\linewidth]{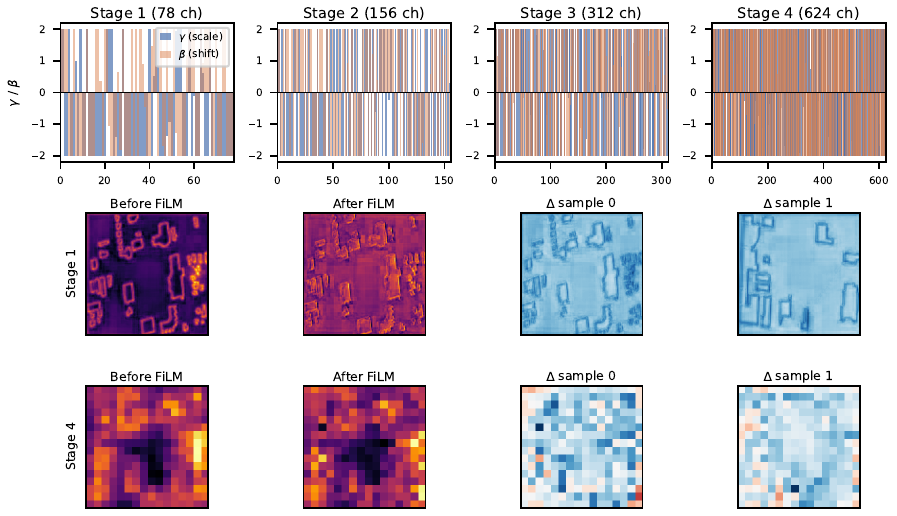}
    \caption{FiLM conditioning analysis. Top: $\boldsymbol{\gamma}$/$\boldsymbol{\beta}$ distributions per stage showing increasing selectivity at deeper stages. Bottom: before/after FiLM activation maps and per-sample difference---FiLM amplifies main-lobe features and suppresses shadow regions.}
    \label{fig:film_analysis}
\end{figure}

\noindent\textbf{FiLM conditioning effect.}
\cref{fig:film_analysis} visualizes FiLM modulation at each backbone stage.
Early stages show near-uniform $\boldsymbol{\gamma}$/$\boldsymbol{\beta}$ while deeper stages develop selective channel amplification, indicating progressive specialization from texture to propagation semantics.
The difference maps reveal that FiLM amplifies features in the main-lobe direction and suppresses shadow regions, consistent with directional antenna radiation physics.

Quantitatively, we partition outdoor pixels by distance from the transmitter.
Near the antenna ($<$150\,px), our model reduces MAE by 52.4\% (0.103$\to$0.049), compared to 44.1\% in far regions.
This indicates that the transmitter spatial channels and FiLM conditioning provide the strongest benefit where geometric relationships between the antenna and surrounding buildings matter most.
In hotspot regions (top-5\% signal strength), improvement is 24.9\% (0.068$\to$0.051), while non-hotspot regions improve by 47.3\%---suggesting the model substantially reduces background noise while maintaining hotspot fidelity.

\begin{figure}[t]
    \centering
    \includegraphics[width=0.9\linewidth]{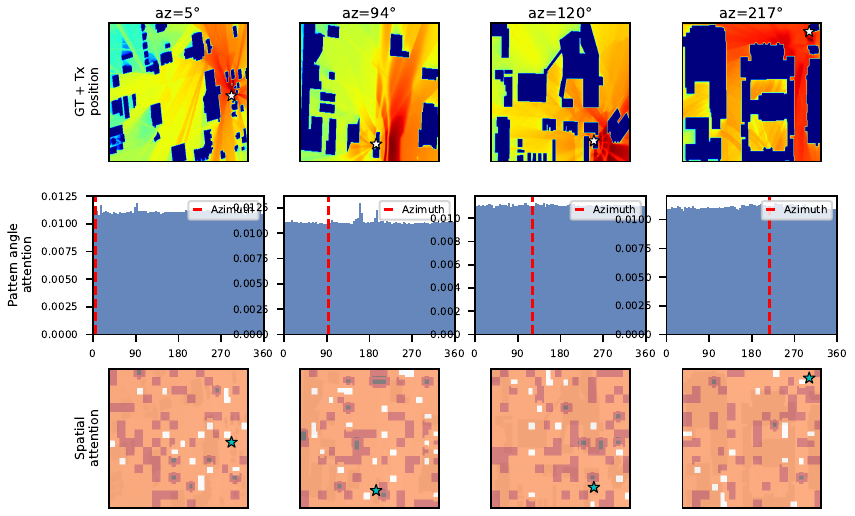}
    \caption{Cross-attention analysis across four antenna azimuths. Top: GT with transmitter (star). Middle: attention weight per pattern angle (dashed = azimuth). Bottom: spatial attention overlaid on GT.}
    \label{fig:cross_attn}
\end{figure}

\cref{fig:cross_attn} shows spatial attention patterns from the cross-attention module across samples with different antenna azimuths.
The spatial maps (bottom) indicate which image regions draw most from the radiation pattern tokens, revealing that the model learns to attend to propagation-relevant areas rather than uniformly distributing attention.

\subsection{Model Complexity}
\label{sec:complexity}

\cref{tab:complexity} summarizes model sizes and inference times.
Our full model (59.3M parameters) is larger than the baselines but provides substantially better accuracy.
The conditioning modules (FiLM + cross-attention) add minimal overhead: comparing the HRFormer+Reg baseline (8.4M, 15.4\,ms) with our full model (59.3M, 19.8\,ms), the increased inference time is primarily due to the larger HRFormer-Base backbone rather than the conditioning modules themselves.

\begin{table}[t]
    \centering
    \caption{Model complexity comparison. Inference measured on RTX 5090 with batch size 1.}
    \label{tab:complexity}
    \begin{tabular}{lccc}
        \toprule
        Model & Parameters & Inference (ms) & Test MAE \\
        \midrule
        Plain UNet & 35.0M & 7.6 & 0.0616 \\
        HRFormer+Reg & 8.4M & 15.4 & 0.0676 \\
        Ours (HRF-Base) & 59.3M & 19.8 & 0.0461 \\
        \bottomrule
    \end{tabular}
\end{table}

% ==============================================================================
% 5. DISCUSSION
% ==============================================================================
\section{Discussion}
\label{sec:discussion}

\noindent\textbf{Why multi-modal conditioning matters.}
The progression from baselines through our ablation rounds reveals that simply increasing model capacity is insufficient---the \emph{way} different modalities are integrated is crucial.
The UNet baseline (35.0M parameters) achieves 0.0616 MAE and 0.326 Hotspot IoU, while our conditioned model (59.3M) reaches 0.0461 MAE and 0.406 IoU.
The spatial error analysis (\cref{sec:analysis}) provides direct evidence: our model improves most near the transmitter (52.4\% MAE reduction) where spatial channels provide explicit geometric priors.

\noindent\textbf{Physics-informed input design.}
The transmitter spatial channels encode fundamental physics priors: signal attenuation increases with distance, and propagation is directionally anisotropic.
Making these priors explicit frees the network to learn complex effects (diffraction, scattering, multi-path) rather than re-discovering basic geometric relationships from data alone.

\noindent\textbf{Limitations.}
Our dataset contains 768 samples from a single city topology at 3.5\,GHz, and we use simulated rather than measured ground truth.
The random split allows the same building layout to appear in training and testing (with different antenna configurations), so the reported metrics reflect within-topology generalization rather than cross-topology transfer.
Generalization to different urban morphologies, frequency bands, and real-world measurements remains to be validated.
The current approach handles 2-D propagation maps and does not model 3-D effects such as vertical signal variation across building floors.
Future work should incorporate larger and more diverse datasets, including real-world measurements, and extend the approach to 3-D prediction.

\FloatBarrier

% ==============================================================================
% 6. CONCLUSION
% ==============================================================================
\section{Conclusion}
\label{sec:conclusion}

We presented a multi-conditioned dense prediction framework for urban EMF map prediction that integrates building layouts, antenna parameters, and radiation patterns through dedicated conditioning pathways.
FiLM modulation injects scalar parameters at every backbone stage, cross-attention fuses angular radiation patterns with spatial features, and transmitter spatial channels provide explicit geometric priors.
A composite loss combining masked L1, MS-SSIM, and focal L1 addresses the prediction difficulty imbalance across EMF maps.
Through controlled experiments, we demonstrate that each component contributes meaningful improvements, and that conditioning mechanism design matters more than backbone scale alone, culminating in a 25--32\% MAE reduction over standard baselines.
Coordinate-consistent TTA provides an additional 6\% improvement by properly handling spatial channel recomputation under geometric transforms.

% ==============================================================================
% ACKNOWLEDGEMENTS
% ==============================================================================
% \section*{Acknowledgements}
% Add acknowledgements here (optional for arXiv).

\bibliographystyle{splncs04}
\bibliography{main}

\end{document}